# Distribution over Beliefs for Memory Bounded Dec-POMDP Planning

**Gabriel Corona**
INRIA
Loria, Campus Scientifique - BP 239
54506 Vandoeuvre-lès-Nancy Cedex
France

**François Charpillet**
INRIA
Loria, Campus Scientifique - BP 239
54506 Vandoeuvre-lès-Nancy Cedex
France

## Abstract

We propose a new point-based method for approximate planning in Dec-POMDP which outperforms the state-of-the-art approaches in terms of solution quality. It uses a heuristic estimation of the prior probability of beliefs to choose a bounded number of policy trees: this choice is formulated as a combinatorial optimisation problem minimising the error induced by pruning.

## 1 Introduction

The problem of the control of a team of cooperative agents in interaction with a stochastic environment has a large field of applications: packet routing in a network, robot coordination, control of sensor networks, zone surveillance etc. The Dec-POMDP (Decentralised Partially Observable Markov Decision Process) formal framework may be used to tackle this problem rigorously. We focus here on the particular case of planning: given the rules of the environment and a goal modelled as a reward function, we are searching for a policy to assign to each agent in order to maximise the rewards. The computation of an optimal joint policy is NEXP-complete in a finite horizon [Bernstein et al., 2000] when the horizon is smaller than the number of states. The exact computation is thus not usable except for very small problems (especially, with a small number of observations) and for small horizons.

Several approaches have been proposed to solve Dec-POMDPs: bottom-up dynamic programming [Hansen et al., 2004], [Szer and Charpillet, 2006], top-down search [Szer et al., 2005] and mathematical programming [Aras et al., 2007]. The number of partial policies to consider in an exact dynamic programming approach is generally exponential in the planning horizon and the number of agents and doubly exponential in the number of observations [Hansen et al., 2004]. Some algorithms have been proposed to overcome this problem: MBDP (Memory Bounded Dynamic Programming) [Seuken and Zilberstein, 2007b], IMBDP (Improved MBDP) [Seuken and Zilberstein, 2007a], MBDP-OC (MBDP with Observation Compression) [Carlin and Zilberstein, 2008], PBIP (Point Based Incremental Pruning) [Dibangoye et al., 2009]. They use only a bounded number of partial plans at each planning step. By bounding the memory, they are able to solve problems with significantly longer horizons than previous algorithms: the complexity is linear with the horizon. By avoiding the exhaustive backup, some of them can solve problems with a higher number of observations.

In this article, we propose a new point based approach, PSMBDP (Policy Search Memory Bounded Dynamic Programming), which uses a heuristic estimation of the prior probability distribution of reachable beliefs and formulates the choice of a bounded number of partial plans as a combinatorial optimisation problem. Using this heuristics information, our algorithm finds solutions of better quality than state-of-the-art approaches such as MBDP and PBIP.

## 2 Models

### 2.1 MDP

The MDP framework (Markov Decision Process) models the decision problem of a single agent which seeks to maximise rewards by interacting with a fully observable environment. The time is discrete and at each time step $t$, the agent knows exactly the state $S^t$ of the environment and chooses an action $A^{t+1}$. A MDP is a tuple $\mathcal{M} = \langle \mathcal{S}, \mathcal{A}, \mathcal{T}, \mathcal{R} \rangle$ where: $\mathcal{S}$ et $\mathcal{A}$ are respectively the sets of states $s$ of the environment and of actions $a$ of the agent; the transition function $\mathcal{T}$ represents the dynamic of the system by defining the transition probabilities, $\mathcal{T}(s'|s,a) = \Pr(S^{t+1} = s'|S^t = s, A^{t+1} = a)$, from state $s$ to state $s'$ when performing action $a$; the reward function $\mathcal{R}$ represents the goal of the agent by defining the expectation $\mathcal{R}(s,a) = \mathrm{E}[R^{t+1}|S^t = s, A^{t+1} = a]$ of the reward $R^{t+1}$ obtained after doing action $a$ in state $s$. In finite horizon $H$, the goal of the agent is to maximise the expected sum of the rewards, $\mathrm{E}[\sum_{t=1}^{H} R^t]$.

### 2.2 POMDP

The POMDP framework (Partially Observable MDP) generalises the MDP framework for partially observable environments: instead of observing the state $S^t$, the agent receives an observations $O^t$ which can be used to infer $S^t$. At each time $t$, the agent chooses an action $A^t$ given the past observations $O^1 \dots O^{t-1}$ and actions $A^1 \dots A^{t-1}$. A POMDP is a tuple $\mathcal{M} = \langle \mathcal{S}, \mathcal{A}, \mathcal{T}, \mathcal{R}, \Omega, \mathcal{O} \rangle$ where: $\Omega$ is the set of observations $o$ that the agent can receive; the observation function $\mathcal{O}$ defines the observation probabilities $\mathcal{O}(o|s,a,s') = \Pr(O^{t+1} = o|S^t = s, A^{t+1} = a, S^{t+1} = s')$ of observing $o$ in the transition from $s$ to $s'$ with action $a$.

A probability distribution $B^t \in \Delta\mathcal{S}$ represents the belief of the agent, i.e. the Bayesian knowledge it has inferred on the current state $S^t$ given the past observations and actions:

$$\begin{aligned} B^t(s) &= \Pr(S^t = s|B^0, A^1, O^1 \dots A^t, O^t) \\ &= \Pr(S^t = s|B^{t-1}, A^t, O^t) \end{aligned}$$

We note $\tau$ the belief update function : $B^t = \tau(B^{t-1}, A^t, O^t)$. The belief is a sufficient statistic for the agent so a POMDP policy is usually a mapping from belief to action.

For convenience, we use the notation:

$$\begin{aligned} \mathcal{O}(o|b,a) &= \Pr(O^{t+1} = o|B^t = b, A^t = a) \\ &= \sum_{(s,s') \in \mathcal{S}^2} \mathcal{T}(s'|s,a)\mathcal{O}(o|s,a,s')b(s) \end{aligned}$$

### 2.3 Dec-POMDP

The Dec-POMDP framework generalises the POMDP framework for a team of agents. A Dec-POMDP is a tuple $\mathcal{M} = \langle \mathcal{I}, \mathcal{S}, (\mathcal{A}_i)_{i \in \mathcal{I}}, \mathcal{T}, \mathcal{R}, (\Omega_i)_{i \in \mathcal{I}}, \mathcal{O} \rangle$ where $\mathcal{I}$ is the finite set of agents $i$. Each agent has its own sets $\mathcal{A}_i$ and $\Omega_i$ of local actions $a_i$ and observations $o_i$: a joint action $a = (a_1, \dots a_n) \in \mathcal{A}$ is a tuple of local actions where $\mathcal{A} = \prod_{i \in \mathcal{I}} \mathcal{A}_i$ is the set of joint actions; a joint observation $o = (o_1, \dots o_n) \in \Omega$ is a tuple of local observations where $\Omega = \prod_{i \in \mathcal{I}} \Omega_i$ is the set of joint observations. At each time $t$, each agent $i$ chooses a local action $A_i^t$ given the past local observations $O_i^1 \dots O_i^{t-1}$ and actions $A_i^1 \dots A_i^{t-1}$. The initial belief $B^0$ is a prior on the initial state $S^0$ shared by the agents. The joint policy $\pi$ is searched as a tuple of local policies $\pi_i$, one for each agent $i$.

The underlying POMDP of a Dec-POMDP $\mathcal{M}$ is the POMDP obtained by replacing the $|\mathcal{I}|$ agents by a single agent receiving the joint observation and choosing the joint action. The underlying MDP is defined as the MDP obtained by observing the state directly.

### 2.4 Policy Trees

In a finite horizon problem, local deterministic policies of a Dec-POMDP can be modelled as policy trees (figure 1) whose nodes are actions and whose branches are observations. We note $a_{q_i}$ the root action of policy tree $q_i$ and $q_i(o_i)$ its subtree for local observation $o_i$. At each time step, the internal state of the agent $Q_i^t$ is defined by a local policy tree $q_i$.

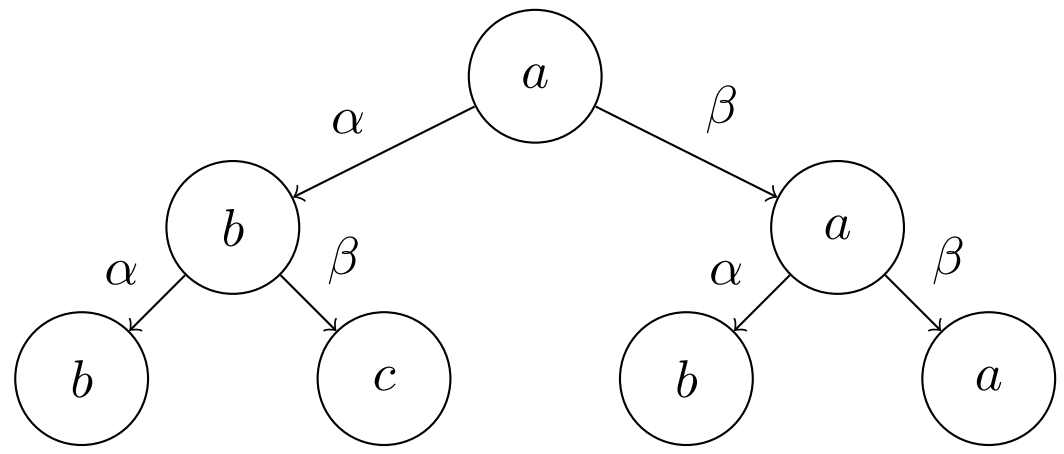

$H = 3$, $\Omega_i = \{\alpha, \beta\}$, $\mathcal{A}_i = \{a, b, c\}$

Figure 1: Deterministic Policy in Finite Horizon

Joint policies trees are tuples of local policy trees (one for each agent), $q = (q_1, \dots, q_n) \in \mathcal{Q} = \prod_i \mathcal{Q}_i$. We note $a_q = (a_{q_1}, \dots, a_{q_n})$ the root joint action of $q$ made of the local root actions $a_{q_i}$ and $q(o) = (q_1(o_1), \dots, q_n(o_n))$ the joint sub-policy tree made of subtrees $q_i(o_i)$ of the local policy trees $q_i$ for the local observations $o_i$. A joint policy tree $q$ can be evaluated using its value function $V_q$ defined for every distribution $b$ on the states:

$$\begin{aligned} V_q(b) &= \sum_{s \in \mathcal{S}} b(s) V_q(s) \\ V_q(s) &= \mathcal{R}(s, a_q) \\ &\quad + \sum_{(s',o) \in \mathcal{S} \times \Omega} \mathcal{T}(s'|s,a_q)\mathcal{O}(o|s,a_q,s')V_{q(o)}(s') \end{aligned}$$

The value function can be represented by its $\alpha$-vector, a $|\mathcal{S}|$-vector containing the $V_q(s)$: $\alpha_q = (V_q(s_1), \dots, V_q(s_{|\mathcal{S}|}))^T$ and $V_q(b) = b \cdot \alpha_q$.

## 3 State of the Art

### 3.1 Dynamic Programming

Dynamic Programming (DP) Dec-POMDP planning builds policy trees in a bottom-up way

[Hansen et al., 2004] (see algorithm 1). The DP operator (line 2) builds the sets $\mathcal{Q}_i^t$ of policy trees of horizon $H-t$ (executed from time $t$ to $H$) from the sets $\mathcal{Q}_i^{t+1}$ of policy trees of horizon $H-t-1$ (executed from time $t+1$ to $H$). This operator is usually made of two phases: backup and pruning (algorithm 2).

**Algorithm 1**: Dynamic Programming planning

**Result**: $\forall i, \mathcal{Q}_i^0$ and their $\alpha$-vectors
1 **foreach** $i \in \mathcal{I}$ **do** $\mathcal{Q}_i^H \leftarrow \{\psi\}$ (empty tree)
2 **for** *t=H-1* **to** *0* **do** $(\mathcal{Q}^t)_{i\in\mathcal{I}} \leftarrow \text{DP}((\mathcal{Q}_i^{t+1})_{i\in\mathcal{I}})$

**Algorithm 2**: General Two Phases DP operator

$(\bar{\mathcal{Q}}_i^t)_{i\in\mathcal{I}} \leftarrow \text{Backup}((\mathcal{Q}_i^{t+1})_{i\in\mathcal{I}})$
$(\mathcal{Q}_i^t)_{i\in\mathcal{I}} \leftarrow \text{Prune}((\bar{\mathcal{Q}}_i^t)_{i\in\mathcal{I}})$

**Backup** The exhaustive backup step builds the sets $\bar{\mathcal{Q}}_i^t$ of local policy trees of horizon $H-t$ by building every tree whose root is an action of $\mathcal{A}_i$ and whose branches associate a local observation to a tree of $\mathcal{Q}_i^{t+1}$ from the previous iteration: $\bar{\mathcal{Q}}_i^t = \mathcal{A}_i \times (\mathcal{Q}_i^{t+1})^{\Omega_i}$. However the number of policy trees of agent $i$ is exponential with the number of local observations $\Omega_i$ and doubly exponential with the horizon.

**Pruning** In order to limit the growth of the number of policy trees, a pruning phase (algorithm 3) is used which consists of eliminating trees of $\bar{\mathcal{Q}}_i^t$ to produce the sets $\mathcal{Q}_i^t$. In exact pruning, the dominated policy trees are pruned. This is checked by solving a linear program. In general, the number of policy trees is still in the same order of magnitude.

**Algorithm 3**: Exact Prune

$\forall i \in \mathcal{I}, \mathcal{Q}_i^t \leftarrow \bar{\mathcal{Q}}_i^t$
flag $\leftarrow$ true
**while** flag **do**
  flag $\leftarrow$ false
  **foreach** $i \in \mathcal{I}$ **do**
    **foreach** $q_i \in \mathcal{Q}_i^t$ **do**
      **if** $q_i$ *dominated* **then**
        $\mathcal{Q}_i^t \leftarrow \mathcal{Q}_i^t \setminus \{q_i\}$
        flag $\leftarrow$ true

### 3.2 MBDP

Memory Bounded Dynamic Programming (MBDP) [Seuken and Zilberstein, 2007b] has been proposed to solve Dec-POMDPs with a high horizon (algorithm 4). It replaces exact pruning by a memory bounded pruning: for each agent $i$, at most $W$ policy trees from $\bar{\mathcal{Q}}_i^t$ are kept in the set $\mathcal{Q}_i^t$. $W$ points (probability distribution on the states) are generated using some heuristic policies. For each point $b$, the best joint policy, $q = (q_1, \ldots, q_n)$, is selected (line 4) and the corresponding local trees $q_i$ are moved from $\bar{\mathcal{Q}}_i^t$ to $\mathcal{Q}_i^t$ (lines 5 and 6).

**Algorithm 4**: DP Operator for MBDP

**Data**: $\mathcal{B}^t$, $W$ points
**Data**: $\forall i, \mathcal{Q}_i^{t+1}$ and their $\alpha$-vectors
**Result**: $\forall i, \mathcal{Q}_i^t$ and their $\alpha$-vectors with $\mathcal{Q}_i^t| \leq W$
1 **foreach** $i \in \mathcal{I}$ **do** $\bar{\mathcal{Q}}_i^t \leftarrow \mathcal{A}_i \times (\mathcal{Q}_i^{t+1})^{\Omega_i}$
2 **foreach** $i \in \mathcal{I}$ **do** $\mathcal{Q}_i^t \leftarrow \emptyset$
3 **foreach** $b \in \mathcal{B}^t$ **do**
4   $(q_i)_{i\in\mathcal{I}} \leftarrow \arg\max_{q\in\prod_i \bar{\mathcal{Q}}_i^t} V_q(b)$
5   **foreach** $i \in \mathcal{I}$ **do** $\mathcal{Q}_i^t \leftarrow \mathcal{Q}_i^t \cup \{q_i\}$
6   **foreach** $i \in \mathcal{I}$ **do** $\bar{\mathcal{Q}}_i^t \leftarrow \bar{\mathcal{Q}}_i^t \setminus \{q_i\}$

**Partial backup** The complexity of the exhaustive backup is exponential with the number of observations. Improved MBDP [Seuken and Zilberstein, 2007a] and MBDP with Observation Compression [Carlin and Zilberstein, 2008], replace the exhaustive backup step by a partial backup in order to scale with the number of observations.

### 3.3 PBIP

Point Based Incremental Pruning (PBIP) [Dibangoye et al., 2009] uses another approach to solve the scalability problem of exhaustive backup. It avoids the exhaustive generation and enumeration of joint policies by using one phase of branch-and-bound search instead of the two phases of backup and pruning (algorithm 5). For each point $b$, a depth-first branch-and-bound search of the optimal joint tree is done (line 5).

**Branch** A search tree node $\tilde{q}$ represents a set of joint policy trees $q$ defined by a set of constraints. It can be seen as a partially defined policy tree. Each child node of the search tree is a subset of its parent defined by an additional constraint on $q$, either by fixing the joint root action $a_q \in \mathcal{A}$ or by fixing a joint subtree $q(o) \in \mathcal{Q}^{t+1}$ for a given joint observation $o$ (see figure 2). The policy tree must not already have been selected: this constraint ($q \notin$ Sel) is checked on the leaf nodes.

**Decentralisability** A joint policy tree $q$ is decentralisable i.e. it can be divided in local policy trees $q_i$ such that:

$$\forall o = (o_1, \ldots, o_n) \in \Omega, q(o) = (q_1(o_1), \ldots, q_n(o_n)) \quad (1)$$

This constraint restricts the valid assignments of subtrees $q(o)$.

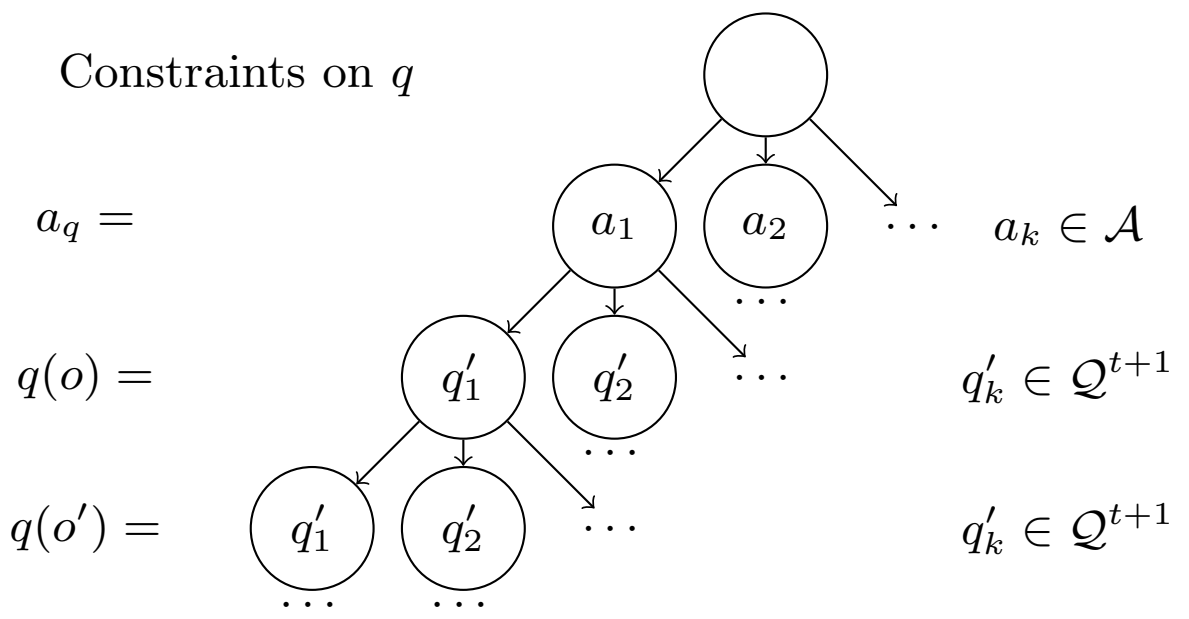

Figure 2: PBIP Search Tree

**Bound** The heuristic value $f(\tilde{q})$ of a search tree node $\tilde{q}$ is a upper bound of the value $V_q(b)$ of the best policy tree $q \in \tilde{q}$ matching the constraints. The set $\Omega$ of joint observations is partitioned in two subsets, $\Omega = \Omega^1 \cup \Omega^2$:

- $\Omega^1$ is the set of joint observations $o$ whose joint subtree $q(o)$ is already fixed, either directly by a $q(o) = q'$ constraint or indirectly through the decentralisability constraint;
- $\Omega^2$ is the set of joint observations $o$ whose joint subtree $q(o)$ is not completely fixed yet; they may be partially constrained by the decentralisability constraint.

The problem is relaxed by ignoring the decentralisability constraints for the subtrees $q(o)$ of observation $o \in \Omega^2$. The contribution of the best joint subtree is used:

$$f(\tilde{q}) = \mathcal{R}(b, a_q) + \sum_{o \in \Omega^1} \mathcal{O}(o|b, a_q) \left[\alpha_{q(o)} \cdot \tau(b, a_q, o)\right] + \sum_{o \in \Omega^2} \mathcal{O}(o|b, a_q) \max_{q' \in \mathcal{Q}^{t+1}} \left[\alpha_{q'} \cdot \tau(b, a_q, o)\right]$$

The theoretical complexity is still exponential with $\max_i |\Omega_i|$ but in practice a large part of the search tree is pruned. In the benchmarks, PBIP handles problems with a higher number of observations than MBDP and finds better solutions than the approaches using partial backup for a lower computation time.

# 4 Proposition

## 4.1 Intuitive Idea

For simplicity of the presentation, our approach is first described in the single agent case. A first approach for the selection of trees consists of maximising the mean of $V^t(b)$ (figure 3):

$$\text{maximise} \int V^t(b)\, db$$

The choice of the set $\mathcal{Q}^t$ of $W$ trees among the $|\mathcal{A}|W^{|\Omega|}$ (generated by exhaustive backup) becomes a combinatorial optimisation problem of variable $\mathcal{Q}^t$:

$$\begin{aligned} \max \quad & \int V^t(b)\, db \text{ with } V^t(b) = \max_{q \in \mathcal{Q}^t} V_q(b) \\ \text{s.t.} \quad & \mathcal{Q}^t \subseteq \bar{\mathcal{Q}}^t = \mathcal{A} \times (\mathcal{Q}^{t+1})^{\Omega} \\ & |\mathcal{Q}^t| \leq W \end{aligned}$$

**Algorithm 5**: DP Operator for PBIP

**Data**: $\mathcal{B}^t$, $W$ points
**Data**: $\forall i, \mathcal{Q}_i^{t+1}$ and their $\alpha$-vectors
**Result**: $\forall i, \mathcal{Q}_i^t$ and their $\alpha$-vectors, with $|\mathcal{Q}_i^t| \leq W$

```
Sel ← ∅
foreach i ∈ I do  Q_i^t ← ∅
foreach  b ∈ B^t do
    Let the search space Q^{t+1} = ∏_i A_i × (Q_i^{t+1})^{Ω_i}
    Search (branch and bound) q = (q_i)_{i∈I} as
    arg max_{q∈Q^{t+1}\Sel} V_q(b)
    Sel ← Sel ∪ {q}
    foreach i ∈ I do  ∀i ∈ I, Q_i^t ← Q_i^t ∪ {q_i}
```

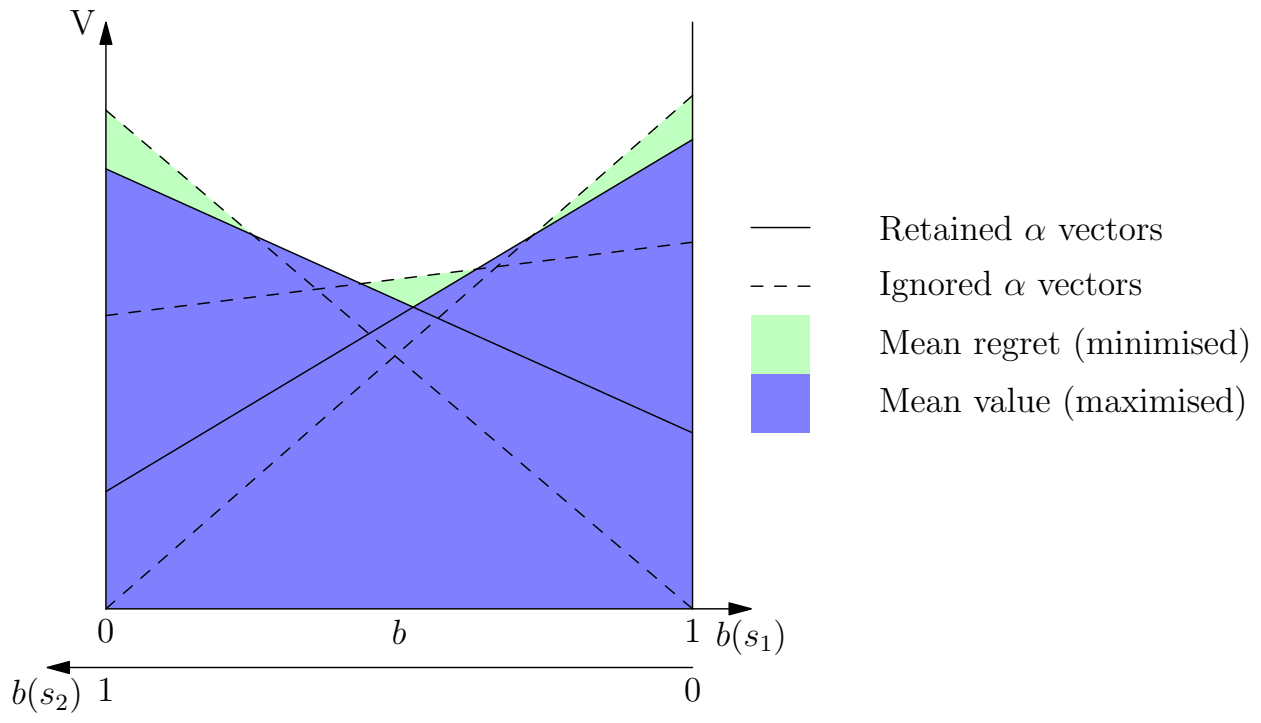

Figure 3: Intuitive Selection Criterion

**Heuristic distribution** Only a subset of the beliefs may be reachable and this information can be used for planning [Szer and Charpillet, 2006]. More generally some beliefs may more probable than others. We introduce a heuristic probability distribution[1] $\mu^t$ of the belief $B^t$ at time step $t$ in the criterion:

$$\text{maximise} \quad \int V^t(b)\mu^t(b)\, db = \mathop{\mathrm{E}}_{b \sim \mu^t}\left[V^t(b)\right]$$

[1] $B^t \in \Delta\mathcal{S}$ is a posterior probability distribution on states $s \in \mathcal{S}$ so $\mu^t \in \Delta\Delta\mathcal{S}$ is a meta-distribution on states.

The prior probability distribution of $B^t$ cannot be determined before planning because it depends on the policy. In practice, a heuristic policy $\tilde{\pi}$ is used to generate the heuristic distributions $\mu^t$. $\mu^t$ is the prior probability distribution[2] of $B^t$ for the given heuristic policy $\tilde{\pi}$: $\mu^t(b) = \Pr(B^t = b | B^0, \tilde{\pi})$.

**Monte-Carlo Sampling** The $\mu^t$ distribution is a finite distribution (in finite horizon, for a finite number of observations and for a given initial belief $B^0 = b^0$) and the integration is a finite sum:

$$\mathop{\mathrm{E}}_{b \sim \mu^t} \left[ V^t(b) \right] = \sum_b \Pr(B^t = b | B^0, \pi) V^t(b)$$

However the size of the set of reachable beliefs is generally exponential with the horizon which hinders exact evaluation. Monte-Carlo sampling on the heuristic probability distribution of $B^t$ can be used to approximate this criterion,

$$\mathop{\mathrm{E}}_{b \sim \mu^t} \left[ V^t(b) \right] \approx \frac{1}{N} \sum_{k=1}^{N} V^t(b_k^t) \tag{2}$$

where the $b_k^t$ are independent samples (the points of the method) of $\mu^t$ and $N$ is the number of samples. Each sample is generated by simulating the heuristic policy until time $t$.

### 4.2 Criterion

A Dec-POMDP can be seen as a POMDP with a constraint on the form of its policy: the policy must be decentralisable. This constraint is added to criterion (2). Given the points $\mathcal{B}^t$, we are searching, for each agent $i$, for a $\mathcal{Q}_i^t$ set of at most $W$ policy trees among the $|\mathcal{A}_i| W^{|\Omega_i|}$ generated by exhaustive backup:

$$\begin{aligned} \text{maximise} \quad & \sum_{b \in \mathcal{B}^t} \max_{q \in \prod_{i \in \mathcal{I}} \mathcal{Q}_i^t} V_q(b) \\ \text{s.t.} \quad & \forall i \in \mathcal{I}, \mathcal{Q}_i^t \subseteq \bar{\mathcal{Q}}_i^t = \mathcal{A}_i \times (\mathcal{Q}_i^{t+1})^{\Omega_i} \\ & \forall i \in \mathcal{I}, |\mathcal{Q}_i^t| \leq W \end{aligned} \tag{3}$$

The samples are taken from the prior distribution of the POMDP beliefs using some heuristic policies. In order to better take into account the decentralised nature of the problem, we might want to work with Dec-POMDP beliefs: it is however not clear how those beliefs should be handled as each agent has its own beliefs; attempts have been made to use them but they are not presented in this paper as they do not provide significant improvements over the first approach on the benchmarks. The heuristics from MBDP can be used as well.

One key difference with MBDP based approaches is that in those approaches each joint policy tree is selected based on only one point $b$. In PSMBDP, the policy trees are selected as a whole based on the distribution $\mu^t$. This distribution is approximated by Monte-Carlo sampling so the number $N$ of points used can be arbitrarily higher than the $W$ parameter.

### 4.3 Algorithm

PSMBDP builds a policy with a memory bounded DP operator: the DP operator uses an exhaustive backup phase and a pruning phase solving (3). Algorithm 6 is a general PSMBDP algorithm using a two step DP operator. Contrary to the approaches based on MBDP, the number $N$ of samples can be higher than the number $W$ of trees which allows to better cover the belief space. In practice, a heuristic solution of (3) is searched.

In the current implementation, criterion (3) is solved using a heuristic greedy approach (algorithm 7). The joint policy tree $q$ maximising this criterion is first selected (line 2): this can be done using a PBIP search for the mean belief, $\bar{b} = \frac{1}{|\mathcal{B}^t|} \sum_{n \in \mathcal{B}^t} b$. Its local trees are placed into the sets $\mathcal{Q}_i^t$ (line 3). Then, the algorithm chooses an agent $i$ and adds the best local tree (searched at line 11) in order to optimise:

$$\max f(q_i) = \sum_{b \in \mathcal{B}^t} \max_{q' \in X} V_{q'}(b) \tag{4}$$

$$\begin{aligned} \text{where } & X = \left\{ (q'_j)_{j \in \mathcal{I}} | q'_i \in \mathcal{Q}_i^t \cup \{q_i\}, \forall j \neq i, q'_j \in \mathcal{Q}_j^t \right\} \\ \text{s.t. } & q_i \in \bar{\mathcal{Q}}_i^t = \mathcal{A}_i \times (\mathcal{Q}_i^{t+1})^{\Omega_i} \end{aligned}$$

This step is repeated until the maximum number of trees is reached for each agent or the criterion cannot be improved further by adding other local trees.

**Algorithm 6**: Two Phases DP Operator for PSMBDP

**Data**: $N$ points $\mathcal{B}^t$
**Data**: $\forall i, \mathcal{Q}_i^{t+1}$ and their $\alpha$-vectors
**Result**: $\forall i, \mathcal{Q}_i^t$ and their $\alpha$-vectors
1 **foreach** $i \in \mathcal{I}$ **do** $\bar{\mathcal{Q}}_i^t \leftarrow \mathcal{A}_i \times (\bar{\mathcal{Q}}_i^{t+1})^{\Omega_i}$
2 $\mathcal{Q}^t \leftarrow \text{Prune}((\bar{\mathcal{Q}}_i^t)_{i \in \mathcal{I}}, \mathcal{B}^t)$, solving (3)

### 4.4 Branch-and-Bound Search

Our original algorithm does not scale with the number of observations because the exhaustive backup builds a number of policy trees exponential with the number of observations. In order to scale better with the number

[2] The $B^t$ random variable is a function of the history of actions an observations, $(\mathcal{A}^1, \Omega^1 \ldots \mathcal{A}^t, \Omega^t)$. We are interested in the problem of prediction of $B^t$ from time 0.

**Algorithm 7**: Prune with Incremental Greedy Optimisation

**Data**: $N$ points $\mathcal{B}^t$
**Data**: $\forall i, \bar{\mathcal{Q}}_i^t$ and their $\alpha$-vectors
**Result**: $\forall i, \mathcal{Q}_i^t$ and their $\alpha$-vectors

1 Let the criterion $\forall \mathcal{Q}, I(\mathcal{Q}) = \sum_{b \in \mathcal{B}^t} \max_{q \in \mathcal{Q}} V_q(b)$
2 $(q_i)_{i \in \mathcal{I}} \leftarrow \arg\max_{q \in \prod_i \bar{\mathcal{Q}}_i^t} I(\{q\})$ (PBIP search)
3 **foreach** $i \in \mathcal{I}$ **do** $\mathcal{Q}_i^t \leftarrow \{q_i\}$
4 **foreach** $i \in \mathcal{I}$ **do** $\bar{\mathcal{Q}_i}^t \leftarrow \bar{\mathcal{Q}_i}^t \setminus \{q_i\}$
5 flag $\leftarrow$ true
6 **while** $\exists i, |\mathcal{Q}_i^t| \neq W \wedge$ flag **do**
7     flag $\leftarrow$ false
8     **foreach** $i \in \mathcal{I}$ **do**
9         **if** $|\mathcal{Q}_i^t| \neq W$ **then**
10             Let $\forall q_i, K(q_i) = (\mathcal{Q}_i^t \cup \{q_i\}) \times \prod_{j \neq i} \mathcal{Q}_j^t$
11             $q_i \leftarrow \arg\max_{\bar{q}_i \in \bar{\mathcal{Q}}_i^t} I(K(\bar{q}_i))$
12             **if** *enhancement of* $I(\mathcal{Q}^t)$ **then**
13                 $\mathcal{Q}_i^t \leftarrow \mathcal{Q}_i^t \cup \{q_i\}$
14                 $\bar{\mathcal{Q}}_i^t \leftarrow \bar{\mathcal{Q}}_i^t \setminus \{q_i\}$
15                 flag $\leftarrow$ true

of observations, the combinatorial optimisation problem (4) (line 11 of algorithm 7) can be solved without explicitly building and exhaustively enumerating the set $\bar{\mathcal{Q}}_i^t = \mathcal{A}_i \times (\mathcal{Q}_i^{t+1})^{\Omega_i}$ of local policy trees: it can be enumerated implicitly by branch-and-bound search similarly to what is done in PBIP. The backup phase is not needed anymore.

**Branch** A search tree node $\tilde{q}_i$ represents a set of local policy trees $q_i$ defined by a set of constraints. It can be seen as a partially defined local policy tree. Each child node $\tilde{q}_i$ of the search tree is a subset of its parent defined by adding a constraint on $q_i$, either by fixing the local action $a_{q_i} \in \mathcal{A}_i$ or by fixing a joint subtree $q_i(o_i) \in \mathcal{Q}_i^{t+1}$ for a given local observation $o_i$ (see figure 4). A difference with PBIP, is that the search is done in the space of local policy trees instead of the space of joint policy trees: the search space is smaller, $|\mathcal{A}_i| W^{|\Omega_i|}$ instead of $|\mathcal{A}|(W^{|\mathcal{I}|})^{\max_{i \in \mathcal{I}} |\Omega_i|}$ for PBIP and every search tree node is valid.

**Bound** The heuristic value $f(\tilde{q}_i)$ of a search tree node $\tilde{q}_i$ is a upper bound of the value $f(q_i)$ for every local policy tree $q_i \in \tilde{q}_i$ matching the constraints (see criterion 4):

$$f(\tilde{q}_i) = \sum_{b \in \mathcal{B}^t} \left( \max_{q \in \mathcal{Q}^t} V_q(b), \max_{q_{-i} \in \mathcal{Q}_{-i}^t} V^+_{\langle \tilde{q}_i, q_{-i} \rangle}(b) \right)$$

where $\tilde{q} = \langle \tilde{q}_i, q_{-i} \rangle = \{\langle q_i, q_{-i} \rangle | q_i \in \tilde{q}_i\}$ is the set of joint policy trees $q = \langle q_i, q_{-i} \rangle$ made of the policy trees

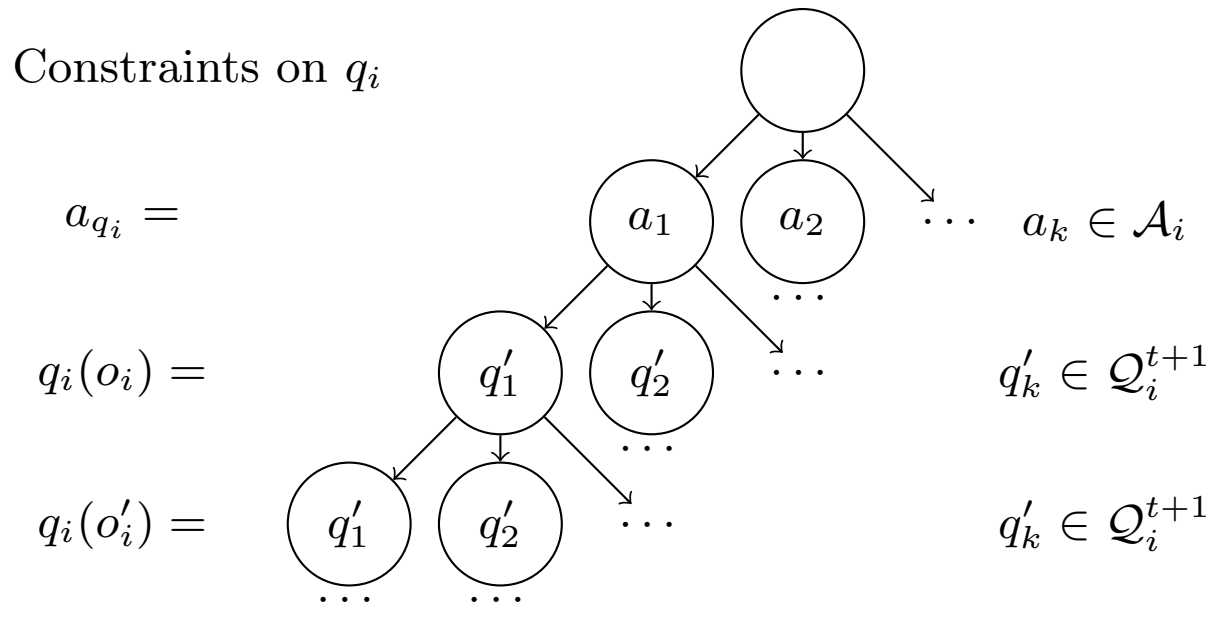

Figure 4: PSMBDP Search Tree

$q_{-i}$ and one local policy tree $q_i \in \tilde{q}_{-i}$. It can be seen as a partially defined joint policy tree made of $q_{-i}$ and the partially defined local policy tree $\tilde{q}_i$. $V^+_{\tilde{q}}$ is a upper bound of the value of the best policy tree $q \in \tilde{q}$. For a given search tree node $\tilde{q}_i$, the set of local observations is partitioned in two subsets, $\Omega = \Omega^1 \cup \Omega^2$: $\Omega_i^1$ is the set of local observations $o_i$ of agent $i$ whose subtree $q_i(o_i)$ is already fixed; $\Omega_i^2$ is the set of local observations $o_i$ of agent $i$ whose subtree $q_i(o_i)$ is not fixed yet. Let $\Omega^1 = \Omega_i^1 \times \Omega_{-i}$ and $\Omega^2 = \Omega_i^2 \times \Omega_{-i}$ be the corresponding sets of joint observations. The problem is relaxed by ignoring the decentralisability constraints for $o \in \Omega^2$ and by choosing a different policy tree $q = \langle q_i, q_{-i} \rangle$ for each belief $b$ and each $q_{-i}$. The contribution of the best possible local subtree $q'_i \in \mathcal{Q}_i^{t+1}$ is used:

$$V^+_{\tilde{q}}(b) = \mathcal{R}(b, a_q) + \sum_{o \in \Omega^1} \mathcal{O}(o|b, a_q) \left[ \alpha_{q(o)} \cdot \tau(b, a_q, o) \right] + \sum_{o \in \Omega^2} \mathcal{O}(o|b, a_q) \max_{q'_i \in \mathcal{Q}_i^{t+1}} \alpha_{\langle q'_i, q_{-i}(o_{-i}) \rangle} \cdot \tau(b, a_q, o)$$

The algorithm is the same as 7 but line 11 is solved with branch-and-bound search and $\bar{\mathcal{Q}}_i^t$ is not computed explicitly by the DP operator.

We have tested both depth-first and best-first search. The latter uses more memory but the memory requirement is still reasonable and the computation is much faster on the tested benchmarks.

As with PBIP, the size of the search space is exponential with the number of observations. If a huge part of the search space is pruned in practice, we have no guarantee that, in general, the explored search space is not exponential with the number of observations.

## 5 Experiments

This section contains results of different algorithms for some benchmark problems, the average expected reward (AEV), its standard deviation ($\sigma$) and the mean computation time ($T$) in seconds.

**Algorithms** PBIP/BeFS is a variation of PBIP with best-first search. PSMBDP and PSMBDP/BeFS are respectively exhaustive PSMBDP and PSMBDP with best-first search. In some cases, the value of the optimal policy of the underlying MDP is shown as well: it is an upper bound of the Dec-POMDP policy values.

**Heuristics** PSMBDP considers the prior of the beliefs of the underlying POMDP when using a heuristic policy and sample belief points on those distributions: a half of the points uses the POMDP policy greedy with respect to the optimal MDP value function; the other half uses the random policy. MBDP and PBIP use MDP heuristic based on the prior distribution on states: a half of the points is taken from the MDP heuristic of MBDP; the other half is taken from the random heuristic of MBDP.

### 5.1 Dec-Tiger

Table 1 shows results for the Dec-Tiger problem [Nair et al., 2003] for $H = 100$, with 50 executions and $N = 100$ for PSMBDP. The (b) variant use the same heuristic as MBDP. Using the POMDP belief prior as heuristic does not give good results. It is not necessary to give a high value of $W$ to PSMBDP to find a good solution because a smaller (usually 5) number of policy trees per time step are kept by PSMBDP.

Table 1: Dec-Tiger Problem

| | $W = 7$ | | |
|---|---|---|---|
| Algorithm | AEV | $\sigma$ | T |
| MBDP | 91.59 | 10.03 | 5.92 |
| PBIP | 88.33 | 10.27 | 0.27 |
| PSMBDP | 91.20 | 3.23 | 0.63 |
| PSMBDP (b) | 166.27 | 7.06 | 1.67 |
| | $W = 20$ | | |
| Algorithm | AEV | $\sigma$ | T |
| MBDP | 138.37 | 3.21 | 540.62 |
| PBIP | 139.68 | 4.59 | 14.84 |
| PSMBDP | Same results as $W = 7$ | | |

Computation times are lower with PSMBDP for this problem, because only a small number of trees per agent are kept: the average width of the local policy trees is 5 for all heuristics. The other trees are not kept as they do not enhance the given criterion. The number of trees and of $\alpha$-vectors generated by exhaustive backup is thus lower.

### 5.2 Firefighters

Results for the firefighting problem [Oliehoek et al., 2008] with 2 agents, 4 houses and 3 fire levels are shown in table 2 for horizon 50 using $W = 7$. For PSMBDP, $N = 100$. PSMBDP finds solutions of better quality than MBDP.

Table 2: Firefighting Problem

| Algorithm | AEV | $\sigma$ | T |
|---|---|---|---|
| Optimal MDP | -35.81 | - | - |
| PBIP | -183.06 | 18.54 | 239 |
| PSMBDP/BeFS | -97.43 | 0.00 | 210 |

### 5.3 Modified Firefighters

In order to test problems with a higher number of observations, we use a modified firefighting problem where each agent observes the resulting fire intensity in the house it goes to. Table 3 show results for this modified problem with $|I| = 2$, 4 houses, fire from 0 to 3, $H = 100$, averaged on 25 executions with $W = 3$ and $N = 100$ for PSMBDP. PSMBDP finds solutions of better quality than PBIP.

Table 3: Modified Firefighting Problem

| Algorithm | AEV | $\sigma$ | T |
|---|---|---|---|
| Optimal MDP | -35.81 | - | - |
| PBIP | -290.93 | 22.28 | 120 |
| PSMBDP/BeFS | -79.39 | 0.00 | 265 |

### 5.4 Cooperative Box Pushing

The Cooperative Box Pushing problem [Seuken and Zilberstein, 2007a] is a problem with a higher number of observations (5 observations per agent for two agents) and has been proposed to benchmark algorithms which can handle a larger number of observations. The results shown in table 4 are average over 25 executions. PBIP times out for $H = 50$ ($T > 20,000s$). PSMBDP finds slightly better solutions than PBIP for a comparable computation time.

## 6 Conclusion

We have introduced a new optimisation criterion to choose the policy trees in a memory-bounded bottom-up dynamic programming approach, i.e. the maximisation of the expected reward given a heuristic policy over beliefs. We have presented PSMBDP, a planning algorithm for Dec-POMDPs based on this principle. A simple implementation using exhaustive enumeration of policies may be used for small problems. In order to handle problems with a higher number of observations, a version with implicit enumeration of joint policy trees by branch-and-bound search may be used.

Table 4: Cooperative Box Pushing

| Algorithm | $W$ | AEV | $\sigma$ | T |
|---|---|---|---|---|
| $H = 20$ | | | | |
| Optimal MDP | - | 511 | - | - |
| PBIP | 7 | 429 | 11.4 | 11,336 |
| PBIP/BeFS | 7 | 421 | 10.3 | 67 |
| | 9 | 432 | 10.2 | 181 |
| PSMBDP/BeFS | 7 | 441 | 11.2 | 160 |
| | 9 | 444 | 12.6 | 189 |
| $H = 50$ | | | | |
| Algorithm | $W$ | AEV | $\sigma$ | T |
| Optimal MDP | - | 1,306 | - | - |
| PBIP | 7 | ? | ? | ? |
| PBIP/BeFS | 7 | 1,063 | 16.2 | 178 |
| | 9 | 1,078 | 13.1 | 471 |
| PSMBDP/BeFS | 7 | 1,076 | 69.7 | 483 |
| | 9 | 1,088 | 31.7 | 590 |

PSMBDP is able to find solutions of better quality than MBDP and PBIP. However, the results depend strongly on the problems and heuristics used. Our optimisation scheme use a simple greedy heuristic so we may be able to find solutions of better quality by using better optimisation methods.